# Abductive Reasoning with the GPT-4 Language Model: Case studies from criminal investigation, medical practice, scientific research

Remo Pareschi[1]

[1]Stake Lab, University of Molise, Campobasso, Italy
remo.pareschi@unimol.it

**Abstract.** This study evaluates the GPT-4 Large Language Model's abductive reasoning in complex fields like medical diagnostics, criminology, and cosmology. Using an interactive interview format, the AI assistant demonstrated reliability in generating and selecting hypotheses. It inferred plausible medical diagnoses based on patient data and provided potential causes and explanations in criminology and cosmology. The results highlight the potential of LLMs in complex problem-solving and the need for further research to maximize their practical applications.

**Keywords:** GPT-4 Language Model, Abductive Reasoning, Medical Diagnostics, Criminology, Cosmology, Hypothesis Generation

## 1 Introduction

The rise of Large Language Models (LLMs) like GPT-4 (OpenAI, 2023) has marked a significant milestone in artificial intelligence, demonstrating an exceptional ability to mimic human-like text. Yet, this progress has sparked intense discussions among scholars. The discourse is largely polarized between two perspectives: one, the critique that these models, often referred to as "stochastic parrots" (Bender et al., 2021)**,** are devoid of true creativity, and two, the counter-argument that they possess an excessive degree of inventiveness often yielding outputs that veer more towards the realm of fantasy than fact. This article investigates these debates, specifically within the context of abductive reasoning, a field that demands a careful balance between creativity and constraint.

Abductive reasoning, often called "inference to the best explanation," involves generating and evaluating hypotheses to explain observations. It is a form of reasoning that is both creative, in that it requires the generation of potential explanations, and constrained, in that these explanations must be evaluated based on their plausibility and consistency with the available evidence. As such, it provides an ideal testing ground to examine the criticisms against LLMs like GPT-4.

In this study, we investigate the abductive reasoning capabilities of GPT-4 in real-world scenarios rather than abstract logical puzzles. This approach is motivated by the belief that the true test of an AI's capabilities lies in its ability to handle the complexity and ambiguity of real-world situations. To this end, we have selected three increasingly complex case studies, each of which requires the use of abductive reasoning.

The first case study concerns a notorious murder case that was solved by a sophisticated DNA analysis of several suspects. The AI assistant's role is to check the logic of the investigation. This case involved generating and testing various hypotheses based on the evidence, offering a suitable context for examining the AI's capacity to comprehend and assess abductive reasoning in a complex, real-world situation.

The second case study is a medical diagnostic scenario, where the AI is tasked with identifying the most plausible explanation for a patient's symptoms. In this case, the AI must decide about the diagnosis and the therapy, even in the face of diagnostic uncertainty. This scenario tests the AI's ability to balance the need for creativity in generating possible explanations and the need for constraint in choosing the most plausible explanation based on the available evidence.

The third and most complex case study involves the exploration of scientific theories in the field of cosmology, specifically theories concerning the origin and fate of the universe. Although partially grounded on empirical data, the speculative nature of these theories lends itself particularly well to abductive reasoning. This case study tests the AI's ability to engage in high-level, abstract reasoning and to generate and evaluate hypotheses in a field where the answers are far from certain.

The article's structure will be as follows: after this introduction, Section 2 will provide background on abductive reasoning, discussing its nature, its role in various fields, and its relevance to the criticisms of LLMs. Section 3 will illustrate our method to test GPT-4's abductive reasoning capabilities. Section 4 will provide three case studies, each in its own subsection, discussing the rationale behind each case and the results obtained. Finally, Section 5 will discuss our findings, their implications for the LLMs debate, and potential avenues for future research.

## 2     Background

Reasoning, the process of thinking about things in a logical, sensible way, is a fundamental aspect of human cognition. It allows us to make sense of the world, make decisions, and solve problems. There are three main types of reasoning: *deductive*, *inductive*, and *abductive*. Each of these types of reasoning has its unique characteristics and applications:

- *Deductive reasoning* is a type of reasoning where the conclusion logically follows from the premises. If the premises are true, the conclusion must be true. It is sound and reliable but lacks creativity as it only applies existing rules to derive conclusions.
- *Inductive reasoning* is a type of reasoning that involves making generalizations based on individual instances. While it is less reliable than deductive reasoning, it is more creative as it allows for generating new hypotheses and theories.
- *Abductive reasoning*, often called "inference to the best explanation," involves generating and evaluating hypotheses to explain observations. It is a form of reasoning that is both creative, in that it requires the generation of potential explanations, and constrained, in that these explanations must be evaluated based on their plausibility and consistency with the available evidence.

Charles Sanders Peirce (1935a, 1935b), an American logician and philosopher, was the first to analyze the three types of reasoning systematically and explicitly defined abduction, which was less familiar and studied than deduction and induction. Reid and Knipping (2019) offer a more recent account. Based on these analyses, as well as insights from other authors and researchers, we can characterize these three types of reasoning by using the following five features:

1. *Creativity* is about generating new ideas, explanations, or solutions. Abductive reasoning is high in creativity, as it involves forming novel hypotheses or explanations for observed phenomena. Inductive reasoning is moderately creative, involving inferring general patterns or principles from specific instances. Deductive reasoning is lower in creativity, as it involves deriving logical consequences from given premises.
2. *Soundness* refers to the logical validity of the reasoning process. Deductive reasoning is high in soundness, as it guarantees the truth of the conclusion if the premises are true and the inference is valid. Inductive and abductive reasoning are lower in soundness as they rely on probabilistic reasoning and hypothesis generation, which do not ensure the truth of the conclusion but only its likelihood or plausibility (Walton, 2004).
3. *Reliability* refers to the dependability of the reasoning process. Deductive reasoning is highly reliable, as it follows strict rules of logic that can be checked and verified. Inductive and abductive reasoning are less reliable as they depend on the quality and quantity of the available evidence, which may be incomplete, inaccurate, or biased (Johnson-Laird, 2008).
4. *Generalizability* refers to the applicability of the reasoning process to different contexts and situations. Inductive reasoning is high in generalizability, as it involves making generalizations from specific instances that can be applied to other cases or domains. Deductive and abductive reasoning are less generalizable, as they involve drawing specific conclusions from given premises or hypotheses that may not hold in other circumstances or settings.
5. *Computability* refers to the ease with which the reasoning process can be implemented in a computational model. Deductive reasoning is highly computable due to its formal, rule-based nature that can be represented and performed by various computational methods, such as search algorithms, knowledge representation languages, and proof systems. Inductive and abductive reasoning are less computable due to their reliance on heuristic processes and the generation and evaluation of hypotheses, which require more human-like intelligence and creativity (Magnani, 2009).

A graphical representation of the distribution of the five features over the three types of reasoning is given in Figure 1.

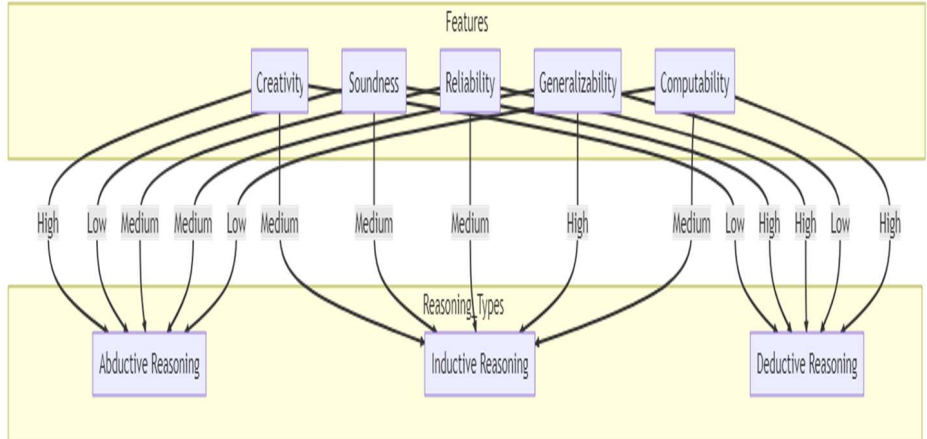

*Figure 1 Reasoning types and their features*

After Peirce introduced abduction, it has been extensively studied and applied in various fields, including artificial intelligence, medicine, and criminal investigation. In the context of this article, some recent works on abduction are of particular interest in that they touch upon aspects relevant to evaluating the performance of LLMs with this type of reasoning. As such, they influenced our way of setting up the interaction with the AI in the test cases. We summarize here some works that were particularly useful in this respect.

Stanley and Nirup's work, "Deductive and Abductive Reasoning with Causal and Evidential Information" (2020), explores the process of medical diagnosis from the perspective of abductive reasoning. They distinguish three aspects of medical diagnosis: generating new diagnostic hypotheses, selecting hypotheses for further pursuit, and evaluating their likelihood in light of the available evidence. They propose conceptualizing medical diagnosis in terms of strategic reasoning, critiquing the threshold approach to clinical decision-making based on decision-theoretic models. Their work provides a detailed exploration of the role of abductive reasoning in medical diagnosis, emphasizing the importance of strategic decision-making in generating and selecting diagnostic hypotheses.

Several papers have examined abductive reasoning in the context of scientific research, highlighting its role in generating creative explanations and models. In his paper, "Abductive Inference: From philosophical analysis to neural mechanisms," Paul Thagard (2007) provides a comprehensive overview of abductive reasoning. He emphasizes its role in generating creative explanations in various domains, including science, medicine, and social behavior. Thagard argues that the essence of abductive reasoning lies in its ability to provide causal accounts, which form the core of explanations. Thagard also discusses the neurological processes enabling minds to make abductive inferences, underscoring the multimodal nature of targets and hypotheses. In "Modeling as Scientific Reasoning—The Role of Abductive Reasoning for Modeling Competence," Upmeier zu Belzen, Engelschalt, and Krüger (2021) claim that abductive reasoning, which involves forming and evaluating explanatory hypotheses, is often neglected in scientific reasoning, while deductive approaches are widely acknowledged. They suggest incorporating abductive reasoning into a framework for modeling competence, arguing that abductive reasoning can be seen as knowledge expansion in model construction. They support their claims with recent studies and examples of learners' performance in modeling processes. The paper "Find out a new method to study abductive reasoning in empirical research" by Żelechowska and colleagued (2020) introduces a novel method for analyzing abductive reasoning in empirical research, combining qualitative and quantitative approaches. The method effectively identifies and analyzes various types of abductive reasoning, capturing its dynamic and complex nature. The study underscores the pivotal role of abductive reasoning in all stages of empirical research, from question formulation to interpretation of findings.

## 3 Evaluation Approach: A Shift from Standard Metrics to Cognitive Assessment

In alignment with the methodology presented in "Sparks of Artificial General Intelligence: Early Experiments with GPT-4" (Bubeck et al., 2023), we diverge from conventional methods of evaluating machine learning models. Traditional metrics, which often concentrate on specific tasks and quantifiable performance measures, may not fully encapsulate a model's capabilities, particularly in complex cognitive tasks involving reasoning, problem-solving, and understanding contexts. Instead, we adopt an approach reminiscent of cognitive psychology's assessment of human cognitive abilities. This method allows for a more holistic and nuanced evaluation of the model's performance, considering its ability to reason, generate hypotheses, and comprehend complex scenarios.

However, our approach also deviates from that of Bubeck et al. (2023) in a crucial aspect. While their methodology emphasizes understanding progression and model enhancement to measure intelligence growth, it remains task-oriented. We favor a dialogical approach for several reasons. Firstly, it aligns with the established methodology of Knowledge Elicitation (KE), developed in the 1980s during the advent of Expert Systems. Based on interview techniques to extract knowledge from domain experts (Hoffman et al., 1995), KE is still applicable today for verifying knowledge stored in LLMs like GPT-4. By adopting KE techniques, we assess performance through in-depth interviews on selected topics, providing a more comprehensive assessment of the model's capabilities.

Moreover, Abduction, the form of reasoning we focus on, naturally fits into a dialogical model where one dialoguer provides an explanation, and the other challenges it (Walton, 2004). Lastly, our interview approach aligns with a vision of human-AI interaction as a collaborative endeavor. This perspective envisions augmented intelligence, where human and artificial intelligence integrate seamlessly, enhancing effectiveness and creativity. This contrasts with a dystopian view where AI replaces human intelligence. Instead, we advocate for a symbiotic relationship, leveraging the strengths of both human and artificial intelligence.

## 4    Case studies

We delve into three case studies where an AI assistant given by the ChatGPT AI assistant, based on the language model GPT-4, has applied abductive reasoning. The results highlight considerable effectiveness and versatility. All case studies were carried out in the form of interviews[1].

**Case Study 1: The Criminal Case of Yara Gambirasio**, In the criminal case of the murder of Yara Gambirasio, which led to the conviction of Massimo Giuseppe Bossetti, the AI assistant used abductive reasoning to verify and reconstruct the investigative process carried out by human investigators. The AI assistant demonstrated its capability to correctly reconstruct the abductive reasoning that led to the identification of the murderer. This case study exemplifies how AI, through abductive reasoning, can be used effectively in criminal investigations, even in complex cases that require a high level of sophistication in reasoning.

---

[1]    The complete interviews can be found at https://osf.io/gz2dk/?view_only=cdbc25bdaee346d498a13fd2114fb9b2

**Case Study 2: Medical Diagnostics.** In medical diagnostics, the AI assistant demonstrated the power of abductive reasoning by inferring the most likely diagnosis based on a patient's symptoms and medical history. In a situation of diagnostic uncertainty, the AI assistant had to make a difficult decision to deal with a young man arriving at the emergency room with progressive symptoms of paralysis. The AI assistant proposed two plausible hypotheses for the patient's condition: botulism and Guillain-Barré Syndrome (GBS) or its variant, Miller-Fisher Syndrome (MFS). The AI assistant's reasoning, diagnosis, and therapeutic indications correspond exactly to the decisions of a medical team that had to deal with the case described in real life. This case study underscores the potential of AI in medical diagnostics, particularly in complex cases that require nuanced and sophisticated reasoning. Given the speed with which the answers were given, it can be of considerable support for real-time decision-making, in the perspective of a collaboration between humans and artificial intelligence in the medical domain, as investigated in (Grasso and Pareschi eds., 2022).

**Case Study 3: Investigation on the Fate of the Universe.** In cosmology, the AI assistant applied abductive reasoning to predict future scenarios based on current observations. It employed a scientific and creative approach to organizing various theories, making explicit the abductive reasoning underpinning them and evaluating each explanation's plausibility. Theories such as the Big Rip, Big Freeze, Big Bounce, or Big Crunch explain the universe's current and potential future states. They are grounded in current observations, like the universe's accelerating expansion, and predict future phenomena.

Unlike the previous cases where the model's performance could be assessed using finite and somewhat resolved scenarios — such as verifying and reconstructing human problem-solving in the Gambirasio crime case or replicating or diverging from human problem-solving in complex medical diagnostic cases — we had to adopt different evaluation methods in this context. Here, the assessment was based on expert evaluations (by researchers in cosmology and astrophysics) of the model's ability to reason within the broad context of hypotheses about the universe's fate. The model's performance in this regard was deemed impeccable.

In this extended dialogue, we could delve deeper into the AI's capabilities. This included its ability to formulate and respond to questions about abduction, identify cause and effect relationships, use deductive logic to validate abductive explanations, and navigate the interplay between deductive, inductive, and abductive reasoning. The AI also showcased its capacity to handle theories that use future predictions to explain both present and future phenomena. This case study comprehensively explores the potential and challenges of employing abductive reasoning in a scientific context. It also highlights the AI's meta-reasoning on abduction, which provides self-awareness about the assumptions used during hypothesis formulation and evaluation. This capability, which the AI demonstrated effectively, can be a valuable contribution to a research team. Moreover, ChatGPT proactively and courteously corrected its interlocutor's mistake regarding the timeline of the decline of one of the theories discussed (the Steady State Theory), showcasing its active engagement in the conversation. Lastly, the AI conducted a self-assessment. The dialogue was conducted with a less powerful version of the assistant, ChatGPT-3.5, represented by a green bar during the interaction, which nonetheless performed effectively. The dialogue was subsequently evaluated by the more advanced version, ChatGPT-4, represented by a violet bar and used directly for the other two interviews.

In summary, these case studies underscore the proficiency of LLM-based AI in employing abductive reasoning across diverse fields and contexts. Whether it's unraveling a criminal case, diagnosing a medical condition, or forecasting the universe's fate, LLMs serve as potent instruments for formulating and assessing hypotheses, even in intricate and sophisticated scenarios. They exhibit rationally bounded creativity, transcending the simplistic label of "stochastic parrots," yet avoiding the pitfall of unrestrained and fantastical inventiveness.

## 5    Conclusion

This preliminary investigation into the capabilities of large language models (LLMs), focusing on GPT-4, has provided encouraging insights into their potential for real-world applications of abductive reasoning. The case studies, which span the domains of criminal investigations, medical diagnostics, and cosmological theories, have demonstrated the model's promising ability to formulate and assess hypotheses in intricate situations.

This study, however, is only the starting point of a larger research effort. To evaluate the model's performance more comprehensively, we need to complement the qualitative approach with a quantitative approach on a larger number of cases. This would involve creating grids of questions to be presented to experts and stakeholders in the fields investigated. This methodical approach would facilitate a comprehensive understanding of the model's strengths and weaknesses in the context of abductive reasoning.

The results thus far are promising, indicating the robustness of LLMs, particularly GPT-4. The model has exhibited a notable level of sophistication in reasoning, demonstrating its potential to tackle complex, real-world scenarios.

Furthermore, this study highlights the potential for a symbiotic relationship between AI and humans. The interactive format of the case studies underscores the value of this collaboration, with a dynamic exchange of feedback between human agents and AI. The AI assistant does not supplant human expertise but enhances it, offering valuable insights and perspectives that may not be readily apparent. This collaborative approach can lead to more effective problem-solving and decision-making processes.

In conclusion, while this research is in its early stages, the initial results are promising. They suggest that with continued research and refinement, AI models like GPT-4 could assume an increasingly significant role in fields where abductive reasoning is paramount. Thus, while this is just the beginning of a journey, we can be optimistic about a future where AI can collaborate with humans in the many contexts where abductive reasoning is required.

**Acknowledgments**

I am grateful to Hervé Gallaire for helpful comments on the general themes of this paper and to Luigi Guzzo and Giovanni Pareschi for providing feedback on the case study on cosmology. The comments from the editors of the monographic issue allowed for further improvements.